# The Energy Worker Profiler: from Technologies to Skills to Realize Energy Efficiency in Manufacturing

*Fareri S[a], Apreda R.[a], Mulas V.[a], Alonso R.[b]*

[a] *Erre Quadro s.r.l*
[b] *R2M Solution s.r.l*

## Abstract

In recent years, the manufacturing sector has been responsible for nearly 55 percent of total energy consumption, inducing a major impact on the global ecosystem. Although stricter regulations, restrictions on heavy manufacturing and technological advances are increasing its sustainability, zero-emission and fuel-efficient manufacturing is still considered a utopian target. For those reasons, energy efficiency is widely seen as a driver of success and coveted at every stage of the production process.

In parallel, the Fourth Industrial Revolution is concretely affecting the way we work and live: companies that have invested in digital innovation over the past 5 years now need to align their internal competencies to maximize their return on investment. More specifically, a primary feature of Industry 4.0 is the digitization of production processes, which offers, among many other things, the opportunity to optimize energy consumption. However, given the speed and often unpredictability with which innovation manifests itself, tools capable of measuring the impact that technology is having on digital and green professions and skills are still being designed.

In light of the above, in this article we present the Energy Worker Profiler, a software designed to map the skills currently possessed by workers, identifying misalignment with those they should ideally possess to meet the renewed demands that digital innovation and environmental preservation impose.

The creation of the Worker Profiler consists of two steps: first, the authors inferred the key technologies and skills for the area of interest, isolating those with markedly increasing patent trends and identifying green and digital enabling skills and occupations. Thus, the software was designed and implemented at the user-interface level. The output of the self-assessment is the definition of the missing digital and green skills and the job roles closest to the starting one in terms of current skills; both the results enable the definition of a customized retraining strategy. The tool has shown evidence of being user-friendly, effective in identifying skills gaps and easily adaptable to other contexts.

# 1. Introduction

The digital revolution is introducing epochal changes and challenges. Companies that have invested in digital innovation in the last years are now in the need for an alignment of their internal competencies to maximize the return on investments; governments are trying to look to the future of sectors that characterize modern economy; universities are reshaping their offer almost every year. In view of this, the digitization has encouraged the conversion of human resource management to an ever more data driven activity, labeled as people analytics (Green, 2017; Kane, 2015), work-force analytics (Hota & Ghosh, 2013; Simon & Ferreiro, 2018), human resources analytics (Lawler, Levenson, & Boudreau, 2004; Levenson, 2005; Rasmussen & Ulrich, 2015), and human capital analytics (Andersen, 2017; Minbaeva, 2017, 2018; Levenson & Fink, 2017; Schiemann, Seibert, & Blankenship, 2017). Beyond the many nomenclatures, what is certain is the interdisciplinary nature of analytics applications, used to analyze performance, engagement and career paths (Bersin et al., 2016). Among these, the topic of skills and job profiles foresight is ever more assuming a key role (WEF, 2016). In fact, the literature highlights not only the necessary integration of existing skills in professional profiles, but also the inevitable creation of new competences and jobs (Galati & Bigliardi, 2019) since the technological requirements demand specific skills to properly manage the digitalisation trends (Castelo Branco et al., 2019; Fareri et al., 2017; Fantoni et al., 2018; Frey & Osborne, 2017). Thus, rapid technological trends calls for frequent updates of jobs and skill taxonomies (ESCO[1], O*NET[2]): even if they embody the state-of-the-art of professions and competencies, they are affected by several limitations (Handel M.J., 2016) and their set of information struggles to stay aligned with the renewed requirements driven by digitization and should be integrated with external findings (Chiarello F. et al., 2021; Malandri L. et al., 2021).

In this regard, the technical knowledge contained in documentations such as papers and patents remains widely unconsidered when trying to predict competence demand, even if results are promising (Karwehl L.J. et al., 2023). Moreover, patent literature contains 80% of the available technical information which cannot be found elsewhere (Geert, A., 2017) and there exists a direct correlation between patents and the impact of innovation, both for the predicting powers that they have (Daim T.U. et al., 2004; Campbell R. S., 1983; Ernst, H., 1997; Liu S.-J. & Shyu J., 1997; Basberg, B.L., 1987) and the level of maturity of the technologies retrievable.

In parallel, in recent years the Industry sector was responsible for almost 55% of the total global energy use, causing an ever more relevant global impact (IEA, 2016). Thus, Energy efficiency is increasingly considered a crucial theme and needs to be managed at all stages of the manufacturing process. Moreover, the primary characteristic of Industry 4.0 is the digitisation of operations, which offers opportunities for energy saving and sustainability (Wachnik B., et al., 2022; Oesterreich & Teuteberg, 2016) through the optimisation of or replacement of technologies, the application of ICT systems for energy efficiency management (Grabot, B. & Schlegel, T., 2014) or adaptation in the business processes.

---

[1] Multilingual classification of Skills, Competences, Qualifications, and Occupations created by the European Commission. https://ec.europa.eu/esco/portal/home
[2] Occupational Information Network

The state of the art seems to focus on building models to assess the digital maturity of companies, considering instead the impact on the labor market as a hazy issue: frequently qualitative approaches are found, while quantitative skill assessments are less explored (Fareri et al., 2020). Notwithstanding, there is a growing skills gap among workers in the manufacturing industry, leading to a lack of skills to meet the needs for rapid technological development and sustainability conversion (Braun, G., et al., 2022; Bukhvald, E.M., et al., 2021).

Given the above, the objective of this research is developing a tool capable of detecting the missing digital and green skills[3] of workers, starting from a self-assessment of their current skill-set: the Energy Worker Profiler. For achieving the goal, the authors started by analyzing the trend of innovation of green technologies in patents. Then, the detection of relevant skills that enable the use of the previous technologies was performed linking the ESCO database with patent insights. Once the database of green and digital skills and jobs was defined, the algorithmic logic of the tool was designed and formalized. Finally, the Worker profiler was implemented as a web platform.

The present paper is structured as follows: first of all, a literature review concerning the value of digital and green competences for energy transition together with a framework of the current HR 4.0 solutions is presented; Secondly, the methodology adopted to reach the analysis objective is deeply described. In the end, the Worker Profiler is presented as well as conclusions and further developments.

## 2. Background

### 2.1 The Impact of Digitization on HRM: analytics, methods and tools

As the Industry 4.0 takes shape, the workforce needs to face an increased complexity of their daily tasks and are required to be flexible and to adapt to new (and challenging) working environments (Longo et al., 2017). Moreover, the recognition of people as the most valuable asset to invest on is widely stated in literature (Srimannarayana, 2010; Boudreau, 1998; Du Plessis and De Wet Fourie, 2016; Nienaber and Sewdass, 2016; Chattopadhyay et al., 2017) as well as the competitive advantage that they could bring (Tootell et al., 2009; De Mauro et al., 2018; Boudreau, 1998). For these reasons, there is a growing scientific interest focused on the design and development of models, tools, practices that could help firms to assess the current skills of the workforce, to improve the evaluation of the human capital and to interact in a frictionless way with new technologies (Longo et al., 2017). In particular, the data driven culture conveyed by Industry 4.0, applied to HR departments, helps through HR analytics to achieve operational and strategic objectives (Huselid, 2018). In more detail, Human Resource analytics is defined as "A HR practice enabled by information technology that uses descriptive, visual, and statistical analyses of data related to HR processes, human capital, organizational performance, and external economic benchmarks to establish business impact and enable data-driven decision-making" (Marler and Boudreau, 2017). In other words, HR analytics is a set of methods and technologies that allows addressing several different issues, such as the evaluation of the performance metrics (Boyd & Gessner, 2013), the consciousness of worker well-being (Greasly & Thomas, 2020) and the assessment of their attitudes, behavior, talents (Shan et al., 2017). In particular, the evaluation of performance and competences based on data is fundamental to both increase efficiency (Garcia-Arroyo & Osca, 2019) and taking fact-based insights reducing subjectivity (Sharma & Sharma, 2017). There are multiple definitions that have been adopted in the last 15 years, which tried to define the boundaries of the topic: extensive use of data and predictive models to drive decisions (Davenport & Harris, 2007), logical analysis that uses business data for reasoning (Fitz-en-2010), evidence based approach for managing people (Bassi et al., 2010; Boudreau & Jesuthasan, 2011), descriptive and visual analysis (Marler & Boudreau, 2017) and scientific and systematic methods to gauge the impact of Human capital management practices (Kryscynski et al., 2017; Van den Heuvel, S. & Bondarouk, T., 2017). To summarize, the purposes are many as well as the methods that could be applied for exploiting data. Another important aspect is represented by the tools that have been designed in the past years. Indeed, the impact of new technologies is particularly relevant when considering the number of web-applications and systems that have been developed for HRM. A clear example is represented by the uses of recruitment tools, chatbots, talent management platforms and, most of all, the use of quantitative mechanisms to assess individual capabilities with a more objective approach (Margherita A., 2022, Giabelli A. et al., 2021). To our best knowledge, tools to assess the current green skills of workers have not been developed yet.

---
[3] Attitudes and skills related to a transition to a circular economy

## 2.2 How to face the need of digital competences to ensure energy efficiency

Not only digitization is affecting the way we work and live. In the last decade, climate change, global warming and the increasing quantity of carbon emissions have become an urgent and critical set of issues to be faced (Kerong et.al, 2022) Moreover, there is a growing anthropogenic impact on elements that are considered the pillars of sustainability: economic, social and environmental (Lupi et al., 2022). In order to mitigate the phenomenon, it has become critical to limit energy consumption and improve energy efficiency, especially in manufacturing industries to face global warming, to cope with rising energy prices and to manage the increasing environmental awareness of customers (Bunse et al., 2010; May and Kiritsis, 2017; Liew et al., 2014). Furthermore, the increasing global energy demand together with the cost of fossil fuels and the downward trend in green energy prices have accelerated this transition (Arigliano et al., 2014). This interest is also evidenced by the development of 'Energy Efficiency First' as a key principle of the EU climate action plan (Dolge et al., 2021). Moreover, energy efficiency is considered a strong driver for meeting the 2030 climate targets since the attempts already made have not been enough to offset the impact of economic growth, which continues to drive energy consumption (European Commission, 2020). Moreover, improvements of energy efficiency in the Industrial sector could have a significant positive impact on reducing emissions (IEA, 2016; Bunse et al., 2010).

In this scenario, emerging technologies are essential drivers to move towards a sustainable industrial approach and gain relevant economic benefits (Lyu and Liu, 2021). In particular, Industry 4.0 enabling technologies (e.g, Artificial Intelligence (AI), Big Data analytics, the Internet of Things (IoT), robotics, blockchain and cloud computing) are considered crucial for realizing the transition. Moreover, the acquisition of Information and Communication Technologies (ICT) jointly with the creation of Cyber-Physical System (CPS) allow the implementation of an environmentally friendly production approach (Lupi et al., 2022). The blockchain technology helps reduce costs, limit delivery times and waste, while IoT, can contribute with data related to the transportation and processing of commercial waste, saving costs and time in recycling processes (Dongfanng et al., 2022).

Sìnce technology development both affects and helps energy intensive industries, the topic of new digital and green skills requirements is becoming extremely strategic (Branca T. et al., 2022; Shamzzuzoha A., et al., 2022). Organizations are required to be aligned with sustainability requirements, also acquiring and training people able to develop, manage, deploy IoT and know the potentiality of artificial intelligence to support green projects (Ogbeibu et al., 2022). Indeed, a new perspective of people management, the sustainable HRM, is increasingly relevant and focused on green performance of the workforce (Mircetic et al., 2022). Furthermore, employees play an even more important role as promoters and recipients of human and organizational sustainability (De Stefano et al., 2017). Against this background, workers and their skills become a fundamental requirement for the transition to a green economy (ILO, 2011). More specifically, the relevant skills for sustainability are technical skills, knowledge, values and attitudes to support sustainable social, economic and environmental outcomes in business (Australian Green Skill Agreement, 2009). In particular, digital marketing and networking, business agility and renewable energy integrated into product manufacturing are currently considered to be the most relevant green skills (Manyati and Mutsau, 2020). To conclude, the detection and acquisition of green skills and job profiles has become indispensable.

# 3. Methodology & Results

In this section, we describe the methodology adopted to develop the database supporting the worker profiler and the tool itself. As we stated in the introduction, patents represent an excellent data source since 80% of technical information is not retrievable elsewhere; additionally, patents provide a valuable overview of where innovation investments are focusing on (Daim T.U. et al., 2004). The designed process is based on exploiting the previous dataset, jointly with the International competence database ESCO as the main source of standard skills and professional profiles, through the use of Text Mining techniques.

The process is graphically summarized in Figure 1. Once the patents were analyzed and the main technologies retrieved, they were compared with ESCO skills through a semantic similarity analysis. Once the key competences were detected, the authors collected related job profiles as well, creating the knowledge-base behind the Worker profiler. Finally, the algorithm logic and the technical specification of the tool was defined and tool implemented. The next sections will explain in more detail each of the main activities of the process and the results obtained.

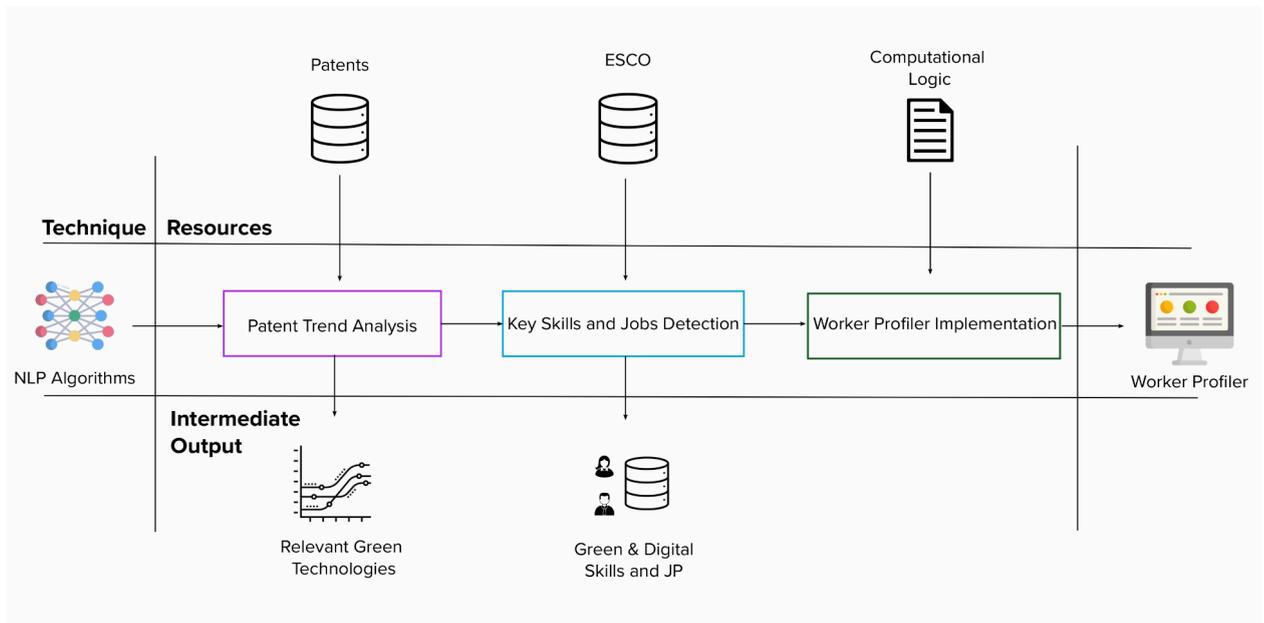

*Figure 1: Workflow of the adopted methodology. The central section shows the main process; the upper section shows the external resources used to carry on the analysis; the bottom section shows valuable intermediate outputs.*

## 3.1 The patent analysis

As reasoned before, patent literature represents a unique source of technical knowledge. They contain information not findable elsewhere, giving an indication on where the innovation is focusing on, ensuring a high level of maturity and feasibility of technologies detectable (higher than those presented in, for example, non-patent articles and literature). The formal structure of patent documents allows the selection of documents focused on specific technologies of interest. In fact, patents describes the technical problem the invention is aimed to solve: this allows the detection of key aspects (characteristics, technologies, solutions, embodiments, components) that are more related to the object of the analysis, avoiding considering false positive signals or taking into account documents where a specific aspect is just mentioned. Furthermore, patents include detailed information on the development of technologies, and forecasting the introduction of a promising technology is a relevant opportunity for companies and countries. By observing the trend of filing patent applications (the number of applications filed over years), together with other parameters (main assignee or the current status of the application), it is possible to understand the maturity level of a technology and estimate its evolution in the next future. In addition, It is possible to extract the list of technologies that allow the resolution of a specific technical problem.

After defining the boundaries of the analysis and building the ontology of keyword to detect all and only the patents of interest, the authors applied text mining tools to extract from the documents phrases containing a technology, also identifying the functions there are related to them and if they are subjects of that function, or objects or if they describe the function itself (verbs). The number of patents related to those chunks over years (the so-called filing trend) were also studied, to define the technology maturity level. The analysis permitted the evaluation of (and how much) those text signals are related to the problem (even weak signals) and, thus, to select only relevant elements. A software aided procedure is then performed, to compare and cluster signals and build the final list of technologies.

To summarize, Patent trend analysis was developed following three steps shown in Figure 2, using the European Patent Office database (Patstat Service) as the main data source, which contains over 120 million patent documents.

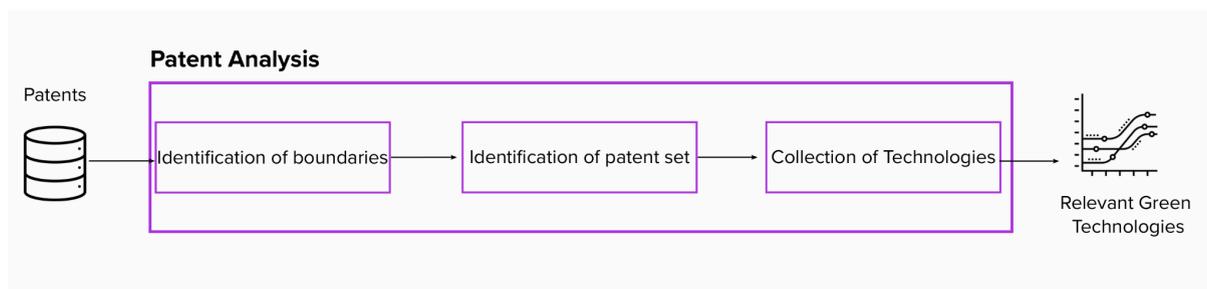

*Figure 2: activities carried out to detect relevant green technologies through patent analysis*

### 3.1.1 Definition of analysis boundaries and Patent set Creation

The boundaries of the analysis were established considering four industrial sectors of interest: Medical Device Manufacturing, Crankshaft Manufacturing (Steelmaking & Forging Processes), Tooling and Mould Production for Appliance, Machining and Production of Composite Products. The latter were combined with the textual content of four different objectives:

1. Digitisation to support sustainable production planning and maximizing the use of renewable energy
2. Reduction of environmental footprint of a crankshaft production by energy-efficient steelmaking and forging processes management
3. Digital Twinning of machining processes to improved planning, design and programming of operations for manufacturing of tooling for Appliance Manufacturing
4. Edge Intelligence for continuous energy optimisation in manufacturing of composite components for industrial machines.

All those elements were propaedeutic to create the ontology of keywords to identify patents focused on energy management systems that could enable the realization of the previous goals. Four different queries, containing keywords and regular expressions related to the previous topics, were drafted and searched in the European Patent Office database (Patstat Service) to define a relevant data set. The patent documents have been managed via semi-automatic tools and algorithms, performing several iterations, to improve both the precision[4] and the recall[5] of each final set. Such iterations are useful to considerably reduce the text ambiguity and reduce the risk of not including documents of interest. The text mining techniques applied, allowed to identify in which section of the text particular keywords or expressions are, their function within the phrase (if they are subject, functions, objects and also nouns, adjectives, verbs) and their correlations with other keywords or expressions. It was thus possible to select documents related to the production processes of interest and primary aspects of the invention itself. Once the document set was selected, it was then possible to retrieve statistical data, to acquire further information (e.g., IPC classes, CPC classes, publications date), evaluate the level of maturity of the technology and compute precision and recall scores, which reached values higher than 90%.

---

[4] It is the fraction of relevant patents among the retrieved patent of the patent set. For the precision evaluation it is possible to perform a selection of a random subset and evaluate the number of relevant documents P' for the field of interest.

[5] It is the fraction of the total amount of relevant patents included in the set.

## 3.1.2 Technologies extraction

Once the patent set was defined, the key technologies (and the phrases in which they were mentioned) needed to be extracted. The term technology has multiple definitions in scientific literature but we choose to adhere to (Puccetti et al., 2023): "a technical mean or in general a technical system created by human-kind through the application of knowledge and science, in order to solve a practical problem or perform a function". Once the field was defined, specific tools of text mining (TM) and a branch of Natural Language processing, Named Entity Recognition (NER), were used. NER systems are widely used to extract general entities (e.g objects, names, cities) but they are also applied to technological (Puccetti et al., 2023) and HR fields (Fareri et al., 2021, Cao et al., 2021). In this work, we applied a rule-based approach (Hearst, 1992), assigning to each term in the patent sentences its role (e.g., noun, adjective) and retrieving hypernym/hyponym relationships using Hearst (1992) fixed patterns. To detect only technologies, we selected hypernym which contains the list of key terms provided by (Puccetti et. al, 2023[6]), enlarged with synonyms obtained through Word2Vec. The technologies were manually revised by authors and clustered using detected hypernym, to obtain a list with a homogeneous level of detail. Finally, statistics were carried out to obtain the trends of the technologies and understand which ones may be emerging and which ones are mature or obsolete. A software aided procedure is then performed, in order to compare and cluster signals, and build the final list of technologies.

## 3.1.3 Results

The patent analysis allowed the identification of a global patent set composed of 22402 patent families and 33 relevant clusters of technologies.
For each technology, the temporal trend was calculated (i.e. the number of patent applications related to each technology filed over years). Based on these results it is possible to evaluate the maturity level of each technology, thus if it is emerging, growing, mature or obsolete. An example of trend is reported in Figure 3, and it is related to Power converter technology. It is possible to observe that there was a peak in 2012 and then the trend is starting to decrease: the main general aspect of the technology has already been investigated, and a new growth is expected related to specific implementations.

---

[6] technology, machine, device, apparatus, mechanism, sensor, network, system, unit.

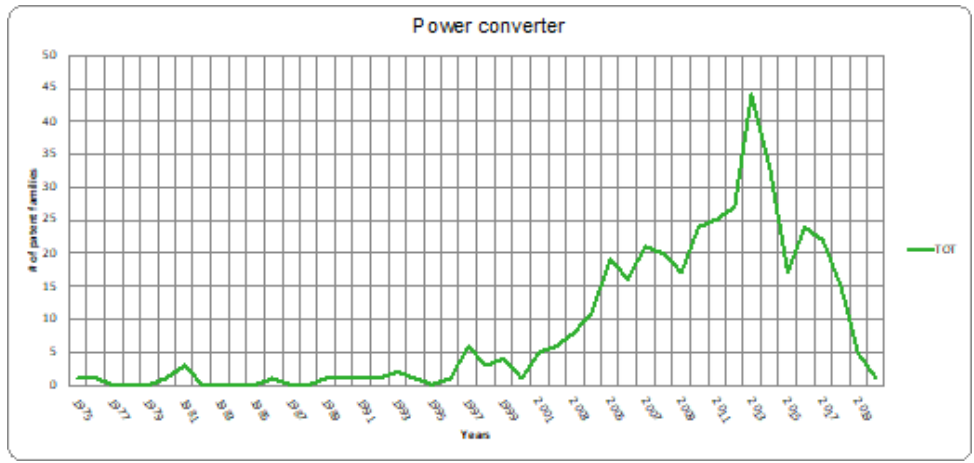

*Figure 3: Trend of Power converter technology.*

What seems evident from our analysis is that, in recent years, companies have focused their attention on energy savings; indeed, almost all of the trends detected in this study showed growing technologies, enabling the transition to the 'smart factory', focused on data collection, communication and management. In general, for some trends it is possible to give more precise information on the technology and therefore to establish whether it is emerging or not, for others this is not possible due the low number of significant patents, which leads to a low statistical consistency. However, almost all the technologies retrieved showed growing trends and it could be explained considering that they are all technologies that are needed to realize the concept of a smart factory. Finally, it has to be considered the field of analysis that we explored: it is also possible that some single technologies are mature if considered by their own, but their application into the field of interest can be new and growing in the next future. To conclude, the list of technology clusters was finally drafted. An extract of them and the % of papers in which they were mentioned is shown in Table 1.

| Technology | % Patents for each technology |
|---|---|
| Air conditioning | 35% |
| Computer device | 33% |
| Heating device | 22% |
| Switching electrical device | 20% |
| Energy storage device | 15% |
| Communication device | 14% |
| Heat exchanger | 11% |
| Light sensor | 11% |
| … | … |

*Table 1: Extract of technologies identified as key to energy efficiency (Source: patents)*

## 3.2 From technology to skill

Starting from the list of key clusters of technologies extracted from the patent analysis and the sentences in which they are contained, it is possible to infer the corresponding enabling competencies, exploiting a valuable data source: ESCO. ESCO is a multilingual system that classifies occupations, skills, competences and qualifications relevant to the labor market in Europe. ESCO contains around 3,000 occupations, the description of around 13,500 skills/knowledge and it provides links to qualifications that certify the acquisition of the competences. The objective of this framework is to provide an overview of the relationship between skills, profiles and qualifications in order to bridge the gap between academia and industry in Europe. The occupation classification corresponds to ISCO-O8, which is the International Standard Classification of Occupations (International Labour Organisation, 2008). To achieve the goal, patent sentences were processed in RStudio Software and compared with ESCO skills/knowledge through semantic similarity algorithm based on BERT[7]; only sentences with a similarity score greater than 0.7 were collected. Semantic similarity analysis was performed through a proprietary algorithm written in Python using PyTorch[8] and Torchvision[9]. For each patent sentence, only the best similarity score with ESCO skill was kept. The list of skills was manually revised by experts to filter duplicate and out-of-scope results. An extract of skills retrieved is shown in table 2.

| Green Skills |
| --- |
| advise on heating systems energy efficiency |
| advise on utility consumption |
| analyze energy consumption |
| design air conditioning |
| design solar energy systems |
| design wind turbines |
| inspect electrical supplies |
| inspect wind turbines |
| install photovoltaic systems |
| … |

*Table 2: Extract of green skills (source: ESCO)*

---

[7] https://en.wikipedia.org/wiki/BERT_(language_model)
[8] https://pytorch.org
[9] https://pytorch.org/vision/stable/index.html

Finally, the job profiles mostly characterized by the previous set of skills were identified, developing the Worker Profiler database. The WP database consists of a list of 87 profiles and more than 300 key competences between hard, soft, green and digital, of which 15% belong to the latter category. An example of key energy efficiency professions, with a description of the purpose of the role, is shown in Table 3.

| Job Title | Job Description |
|---|---|
| **Energy Engineers** | Energy engineers design new, efficient and clean ways to produce, transform and distribute energy in order to improve environmental sustainability and energy efficiency. They derive energy from natural resources, such as oil or gas, or renewable and sustainable sources, such as wind or sun. |
| **ICT managers for Environmental Sustainability** | ICT managers for environmental sustainability know the legal framework for green ICT, understand the role of ICT network configurations in the economy and distribution of energy resources, and assess the carbon footprint impact of each ICT resource in the organisation's network. They plan and manage the implementation of environmental strategies for ICT networks and systems, conducting applied research, developing organisational policies and devising strategies to achieve sustainability goals. They ensure that the entire organisation uses ICT resources in the most environmentally friendly way possible. |
| **Sustainability Managers** | Sustainability managers are responsible for ensuring the sustainability of business processes. They assist in the development and implementation of plans and measures to ensure that production processes and products comply with certain environmental and social responsibility standards and monitor the implementation of sustainability strategies within the company's supply chain and business process and report on the results. They analyse issues concerning production processes, material use, waste reduction, energy efficiency and product traceability in order to improve environmental and social impact and integrate sustainability aspects into the corporate culture. |

*Table 3: Example of key professions detected in international skills databases*

Once the relevant archetypes were defined, the authors proposed a classification into three macro-categories, obtained using both a top-down and a bottom-up approach. Following the top-down approach, the archetypes were labeled according to the macro-category declared in ESCO. Through the bottom-up approach, the authors validate the goodness of the previous result through a cluster analysis (Figure 4) based on the semantic similarity between profiles composing each class. The result of the similarity analysis confirmed the top-down classification and represents a cross-validation process of our categorization. The final three classes of archetypes are Technicians-Operators, Managers-Consultants and Engineering professionals. The definition of the three classes are reported in Appendix 1.

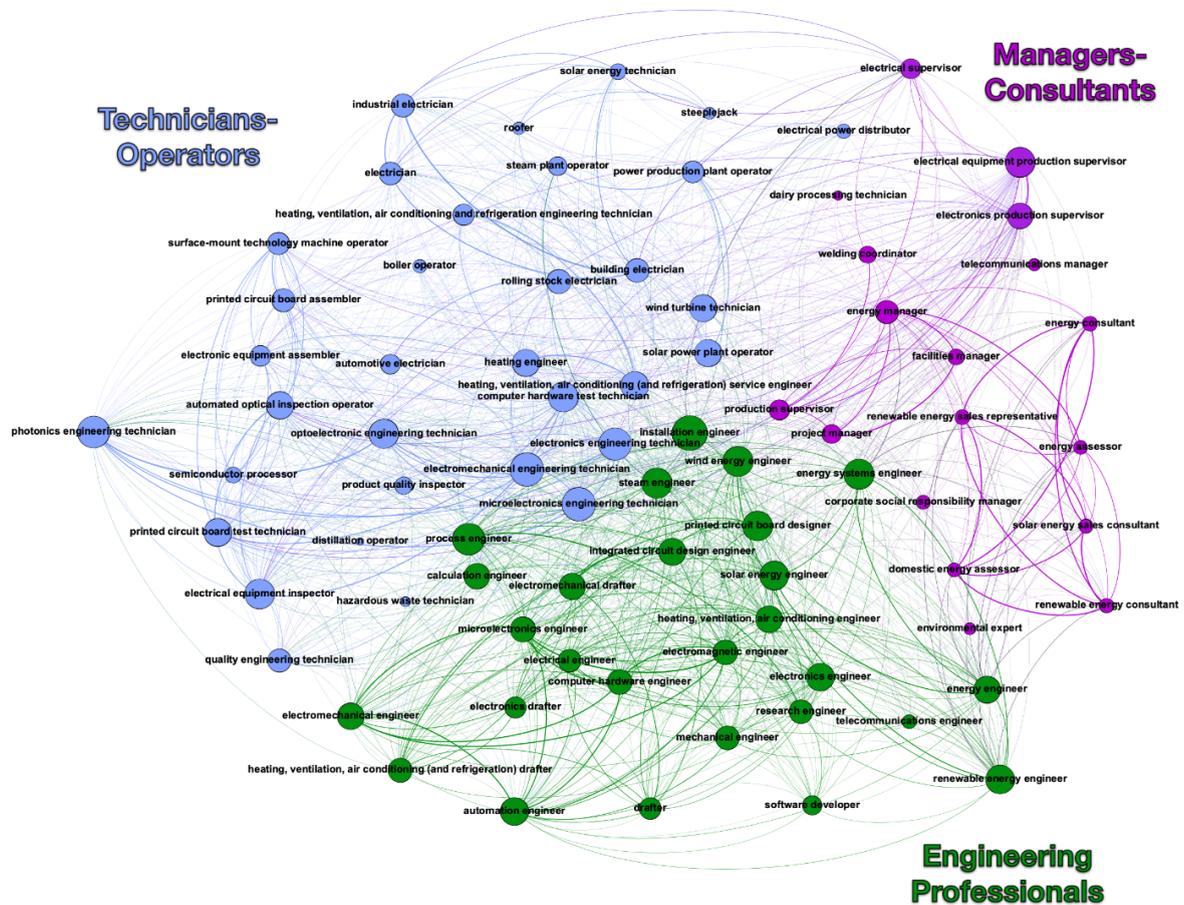

*Figure 4: data-driven approach to detect cluster of jobs*

## 3.3 The Worker Profiler

The Worker Profiler[10] is a tool that enables the user to perform a self-assessment of their own skills, with a particular focus on green skills (and skills gap). The software is also able to identify, starting from the selection of one's profession, the profiles most akin to that of the user in terms of skills currently possessed, in order to highlight (prospective) activities of effective job rotation. The platform was designed and developed following minimalist UI principles, with the objective of having a clean looking, multi-device and multi-platform UI, fully working in PCs but also in Tablets and Mobile devices. For these reasons we relied on responsive Bootstrap code augmented with some ad-hoc CSS code, carefully selected Javascript libraries and a collection of SVG Icons. The application also takes into account the privacy of workers, allowing them to control and manage all their assessments and data.

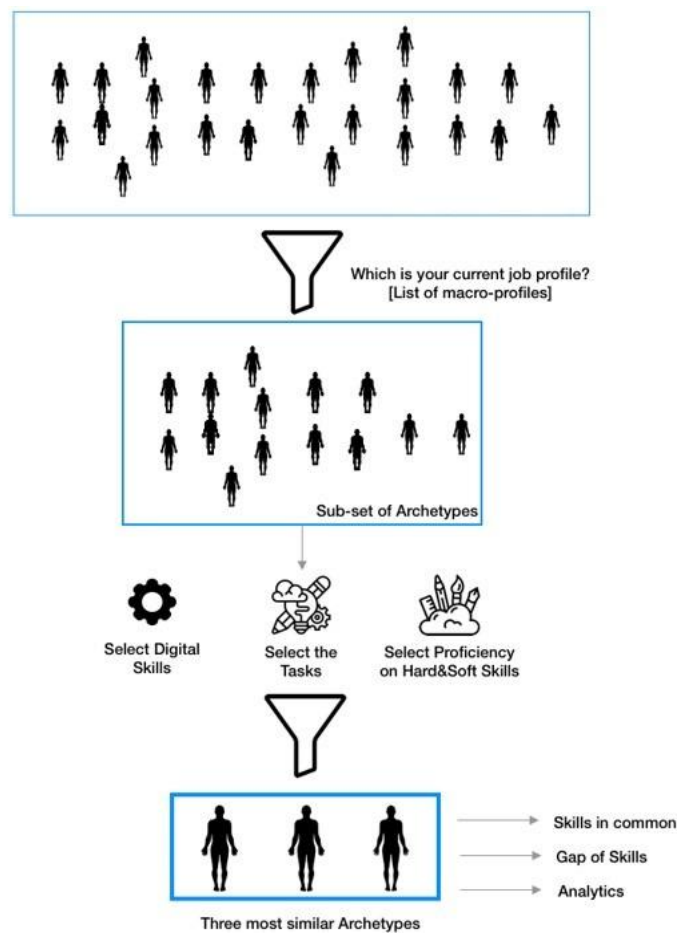

*Figure 5: main steps of the Worker profiler*

---

[10] The tool is available here: https://denim.r2m.cloud/register.html

The tool, which has the appearance of a questionnaire, was built taking the database of skills and professions described above as the main data source. Considering the user-experience (Figure 5), the worker is asked to select the professional profile with which he or she best identifies from the list of 87 professional archetypes. Subsequently, the users will choose the hard skills, digital skills, and soft skills they currently possess. Finally, the tool calculates the distance between the answers and the ideal skill set of the profile of interest, providing also information about soft skills to improve or maintain for the current occupation (Figure 6) and offering the user a detailed assessment of those (hard, digital, green) to be acquired and evidence of the three professional profiles most similar to it. The tool is user-friendly and useful to propose ad hoc-training experience and gives a data-drive indication to realize an effective job rotation.

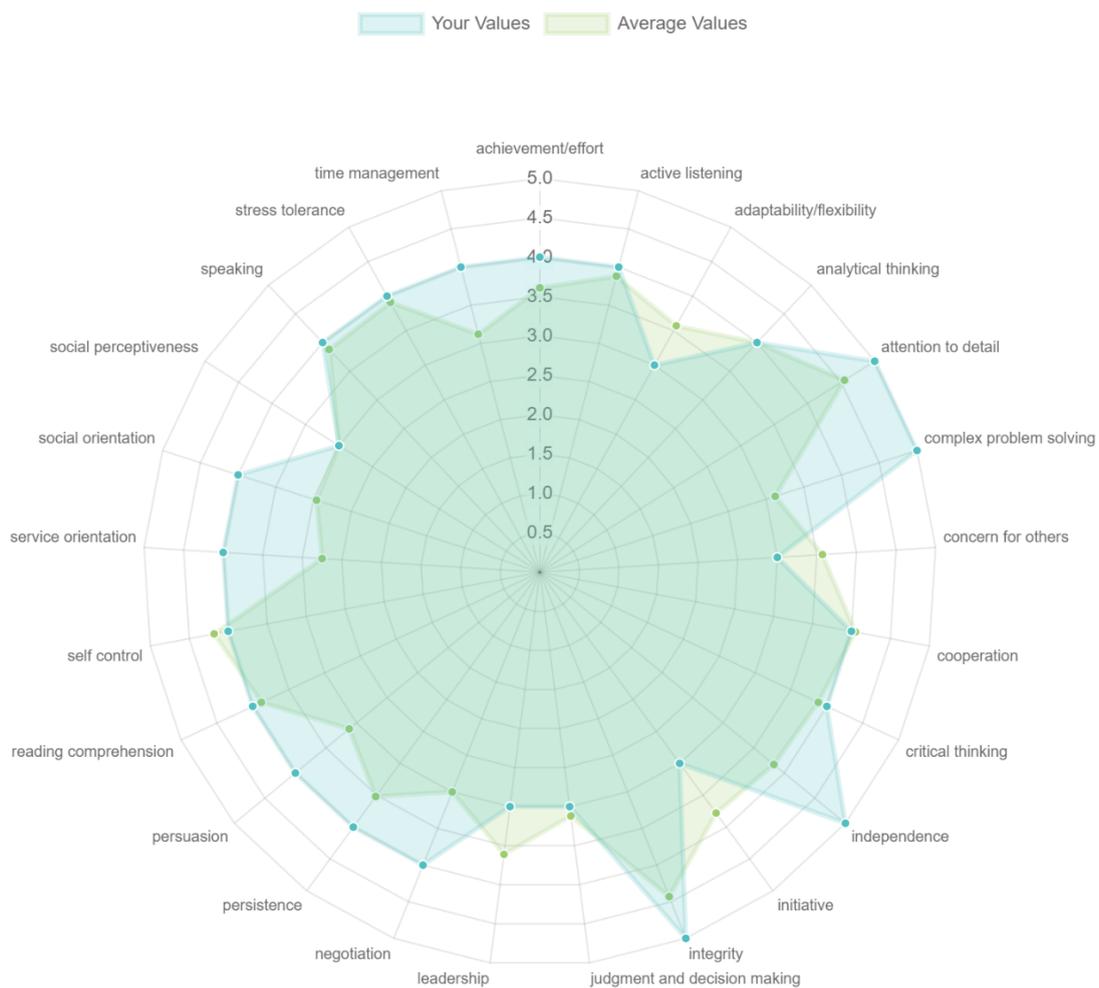

*Figure 6: soft skill comparison chart*

## 4. Conclusions

Due to the increasing relevance of energy efficiency in manufacturing, designing, managing and maintaining production models with reduced environmental impact, fully adopting enabling technologies, and operationalizing investments made in new digital technologies is essential and requires a broad and complex transformation of the skills of the workers. In this context, Artificial Intelligence and tools that rely on data-driven approaches can be useful for all companies engaged in this transformation. Artificial Intelligence helps automate steps and analyses (such as the so-called "skill gap" analysis, i.e., measuring the distance between current and ideal skills) by representing an assessment tool, or self-assessment, within everyone's reach. In this article we present the Energy Worker Profiler, a software designed to map the skills currently possessed by workers, identifying the mismatch with those they should ideally possess in order to meet the renewed demands that digital innovation and environmental protection impose. The Worker Profiler was built starting from the detection of key technologies and skills for the area of interest (in this case, adoption of digital technologies for efficient energy management in industry), identifying those with distinctly increasing trends and identifying green and digital skills and professions that enable them. In more detail, starting from four sectors of interest and four different digitization and sustainability objectives, a patent analysis was conducted to detect the technologies of interest with a marked increasing trend. After that, semantic similarity algorithms were applied to the International Competence Database (ESCO), to detect a set of enabling skills and related job profiles. The number of relevant job profiles retrieved was 87 and the relevant skills more than 300. The knowledge base was then taken as input for the semi-automatic creation of a questionnaire to evaluate the current skills of workers and to assess the key green e digital gaps. After that, a User-interface was developed to make the Worker Profiler accessible.

The output of the self- assessment is the detection of the 3 archetypes the worker embodies the most, and the differences between the skills they currently possess and the ones that they need to acquire. The latter represents the starting point to detect ad hoc training courses and to define the customized up-skilling strategy. To conclude, the assessment process aims to be data-driven, efficient and easy to replicate in other contexts and industrial sectors and will offer evidence-based approaches to enhance the maturity and workforce skills towards digital and sustainable factories of the future.

# Acknowledgments

This study was partly founded by the EU HORIZON 2020 project DENIM (Digital Intelligence for Collaborative Energy management in Manufacturing) Grant Agreement N. 958339.

# Appendix 1: definition of job macro-classes

| **Macro-Class** | **Definition** |
| --- | --- |
| *Technicians-Operators* | Operators & Technicians perform technical tasks connected with research and operational methods in science and engineering. They operate and monitor plants and adjust and maintain processing units and equipment. They supervise and control technical and operational aspects of manufacturing, construction and other engineering operations. |
| *Engineering Professionals* | Engineering professionals design, plan and organize the testing, construction, installation and maintenance of structures, machines and their components, and production systems and plants; and plan production schedules and work procedures to ensure that engineering projects are undertaken safely, efficiently and in a cost-effective manner. |
| *Managers-Consultants* | Managers plan, direct, coordinate and evaluate the overall activities of enterprises, governments and other organizations, or of organizational units within them, and formulate and review their policies, laws, rules and regulations. Consultants provide advice on how to optimise the use of existing tools and systems, make recommendations for the development and implementation of a business project or technological solution and contribute to project definitions. |

*Table A1: Archetypes's macro-classes and their definitions*